\documentclass{article}

\usepackage{microtype}
\usepackage{graphicx}
\usepackage{subfigure}
\usepackage{booktabs} 
\usepackage{amsmath}
\usepackage{float}
\usepackage{amssymb}
\usepackage{dblfloatfix}

\usepackage{hyperref}

\usepackage[accepted]{icml2020}

\icmltitlerunning{Evaluating the Fairness Impact of Differentially Private Synthetic Data}

\begin{document}

\twocolumn[
\icmltitle{Evaluating the Fairness Impact of Differentially Private Synthetic Data}



\icmlsetsymbol{equal}{*}

\begin{icmlauthorlist}
\icmlauthor{Blake Bullwinkel}{iacs}
\icmlauthor{Kristen Grabarz}{iacs}
\icmlauthor{Lily Ke}{iacs}
\icmlauthor{Scarlett Gong}{iacs}
\icmlauthor{Chris Tanner}{iacs}
\icmlauthor{Joshua Allen}{msft}
\end{icmlauthorlist}

\icmlaffiliation{iacs}{IACS, Harvard University, Cambridge, USA}
\icmlaffiliation{msft}{Microsoft, Redmond, Washington, USA}

\icmlcorrespondingauthor{Blake Bullwinkel}{jbullwinkel@fas.harvard.edu}

\icmlkeywords{Machine Learning, ICML}

\vskip 0.3in
]



\printAffiliationsAndNotice{}  

\begin{abstract}

Differentially private (DP) synthetic data is a promising approach to maximizing the utility of data containing sensitive information. Due to the suppression of underrepresented classes that is often required to achieve privacy, however, it may be in conflict with fairness. We evaluate four DP synthesizers and present empirical results indicating that three of these models frequently degrade fairness outcomes on downstream binary classification tasks. We draw a connection between fairness and the proportion of minority groups present in the generated synthetic data, and find that training synthesizers on data that are pre-processed via a multi-label undersampling method can promote more fair outcomes without degrading accuracy.

\end{abstract}

\section{Introduction}

Data containing sensitive information on individuals are being collected in an increasing number of domains. In fields such as healthcare, publishing analyses of sensitive data is not only an important part of the research process but also has clear benefits for humanity. However, publicly releasing these analyses creates a privacy risk for the individuals represented in the underlying data sets. Even machine learning models, including neural networks, that perform complex transformations of their input data can leak information about individual records in their outputs \cite{shokri}.

Differential privacy (DP) has emerged as the gold standard for private data protection and provides strong theoretical guarantees \cite{dwork}. Specifically, differential privacy bounds privacy loss by a pre-specified parameter and ensures that the output of a computation does not reveal personal information that could not be inferred from population-level statistics.

Recent research has focused on DP implementations of machine learning models, many of which are trained using modified versions of optimization algorithms that add noise to gradient computations \cite{abadi}. This approach is well established but has a major drawback: after a model has been trained with a certain privacy loss budget, no further computations can be performed on the outputs. This is particularly problematic when it is necessary to perform multi-step computations on complex data sets. Further, it severely limits the ability of machine learning practitioners to understand their data and share their results.

This drawback motivates differentially private synthetic data, a growing research topic focused on training generative models to produce synthetic data that preserve the statistical properties of the original data. By distributing the privacy loss parameter across noise added to gradients during training, these models satisfy the definition of differential privacy while providing data that can be used for an unlimited number of subsequent computations \cite{rosenblatt}.

Because synthetic data distributions are mere approximations of the original data, privacy guarantees are often granted at the expense of predictive accuracy \cite{geng}. Furthermore, recent work has shown that this expense may not be borne equally across protected classes represented in the data. In other words, differentially private machine learning algorithms may be less likely to satisfy common definitions of fairness \cite{bagdasaryan}. In the context of models trained on differentially private \textit{synthetic} data, however, the interplay between privacy and fairness is not well understood. We investigate this relationship by training binary classification models on DP synthetic data generated by four synthesizers at a range of privacy budgets.

\section{Background and Preliminaries}

The synthesizers we evaluated use different methods to satisfy differential privacy. As described in Definition 1, DP ensures that the output of a computation performed on a data set is statistically indistinguishable from the output of the same computation on the data set with any individual's information removed, up to some privacy loss parameter $\varepsilon$.

\textbf{Definition 1.} (Differential Privacy \cite{dwork}) \textit{A randomized function $f$ provides $(\epsilon,\delta)$-differential privacy if $\forall S \subseteq Range(f)$, all neighboring data sets $D$, $\hat{D}$ differing on a single entry,} 
\begin{equation}
    Pr[f(D) \in S] \leq e^\epsilon Pr[f(\hat{D}) \in S] + \delta
\end{equation}

DP synthesizers allow us to control exactly how much privacy loss, or privacy risk, we are willing to tolerate. Lower values of $\varepsilon$ are associated with greater privacy protection, whereas higher values typically allow greater statistical similarity to the original data, but preserve less privacy.

\textbf{Synthesizers.} In our experiments, we generated data using four DP synthesizers. One of these is the Multiplicative Weights Exponential Mechanism (MWEM), developed by Hardt et al. (\citeyear{hardt2012simple}). Using the Multiplicative Weights update rule and the Exponential Mechanism to select queries, MWEM approximates a target distribution by generating data that maximize agreement with the target on the selected queries. Recent research has found that MWEM is fast and performs well in scenarios where data can be discretized into columns with reasonable dimensionality \cite{rosenblatt}. As MWEM was one of the first DP synthesizers published, we include it as a de facto baseline. 

Another popular approach to private data generation is to train GAN-based models \cite{gans} using DP Stochastic Gradient Descent (DPSGD), which enforces privacy by clipping each gradient in the optimization's $L_2$ norm and then adding noise \cite{abadi}. We evaluated two synthesizers that fall under this category: DP-CTGAN, a DP version of CTGAN developed to generate private tabular data \cite{xu2019modeling}, and PATE-CTGAN, which combines the Private Aggregation of Teacher Ensembles (PATE) framework with CTGAN \cite{yoon2018pategan}. While GAN-based models can generate high quality synthetic data, we found that their performance is sensitive to hyperparameters, making them difficult to train across data sets. To mitigate this issue, we used QUAIL, an ensemble method proposed by \citeauthor{rosenblatt} that combines a DP supervised learning algorithm, such as DP logistic regression, with a synthesizer. In our experiments, we augmented MWEM, DP-CTGAN and PATE-CTGAN with QUAIL and refer to these ensembles as QUAIL-MWEM, QUAIL-DPCTGAN and QUAIL-PATECTGAN, respectively. For these synthesizers, we utilized the open-source \href{https://github.com/opendp/smartnoise-sdk}{SmartNoise SDK}.

Finally, we evaluated MST \cite{mckenna_mst}, a more recent synthesizer that uses the Gaussian mechanism to measure selected marginals and Private-PGM to estimate a distribution from those measurements and generate synthetic data. As a graphical model, Private-PGM tends to work well in high dimensions, provided that the selected marginals are low-dimensional \cite{private_pgm}.

\textbf{Fairness.} In this work, we considered the equalized odds notion of fairness. While many definitions of fairness exist, equalized odds provides a measure that is both relevant to our classification tasks and more robust than other definitions. Demographic parity, for example, does not guarantee fairness in all scenarios and can seriously hurt accuracy when enforced \cite{dwork2011fairness}. As implied by Definition 2, equalized odds requires that the true positive rate ($\hat{Y}=1,y=1$) and false positive rate ($\hat{Y}=1,y=0$) across groups (e.g. gender, race) are equal.

\textbf{Definition 2.} (Equalized Odds) \textit{A classifier $\widehat{Y}$ satisfies equalized odds with respect to a protected attribute $A$ and outcome $Y$ if $\widehat{Y}$ and $A$ are independent, conditional on $Y$,}
\begin{multline}
    Pr[\widehat{Y}=1 | A=0, Y=y] \\ = Pr[\widehat{Y}=1 | A=1, Y=y], \quad y \in \{0,1\}
\end{multline}


Note that in our analyses, we labeled the unprivileged group $A=0$ and the privileged group $A=1$. In all four data sets analyzed, the true positive and false positive rates are aligned with real-world fairness concerns. In the context of the COMPAS data set, for example, a higher false positive rate for group $A=0$ than $A=1$ implies that a classifier is more likely to incorrectly predict that individuals in the unprivileged group will recommit a crime compared to individuals in the privileged group.

To measure the degree of \emph{unfairness} between groups, we calculated the difference between their true positive rates and false positive rates, i.e. $Pr[\widehat{Y}=1 | A=1, Y=y] - Pr[\widehat{Y}=1 | A=0, Y=y]$, where $y \in \{0,1\}$. We refer to these differences as the equalized odds distances, smaller values of which indicate more fair outcomes. Note that in addition to fairness, we also measured accuracy using F1-scores because a classifier that always predicts $\hat{Y}=0$ would perfectly satisfy equalized odds, thereby appearing very fair but offering no predictive value.

\begin{figure*}[!h]
\centering
\includegraphics[width=0.7\textwidth]{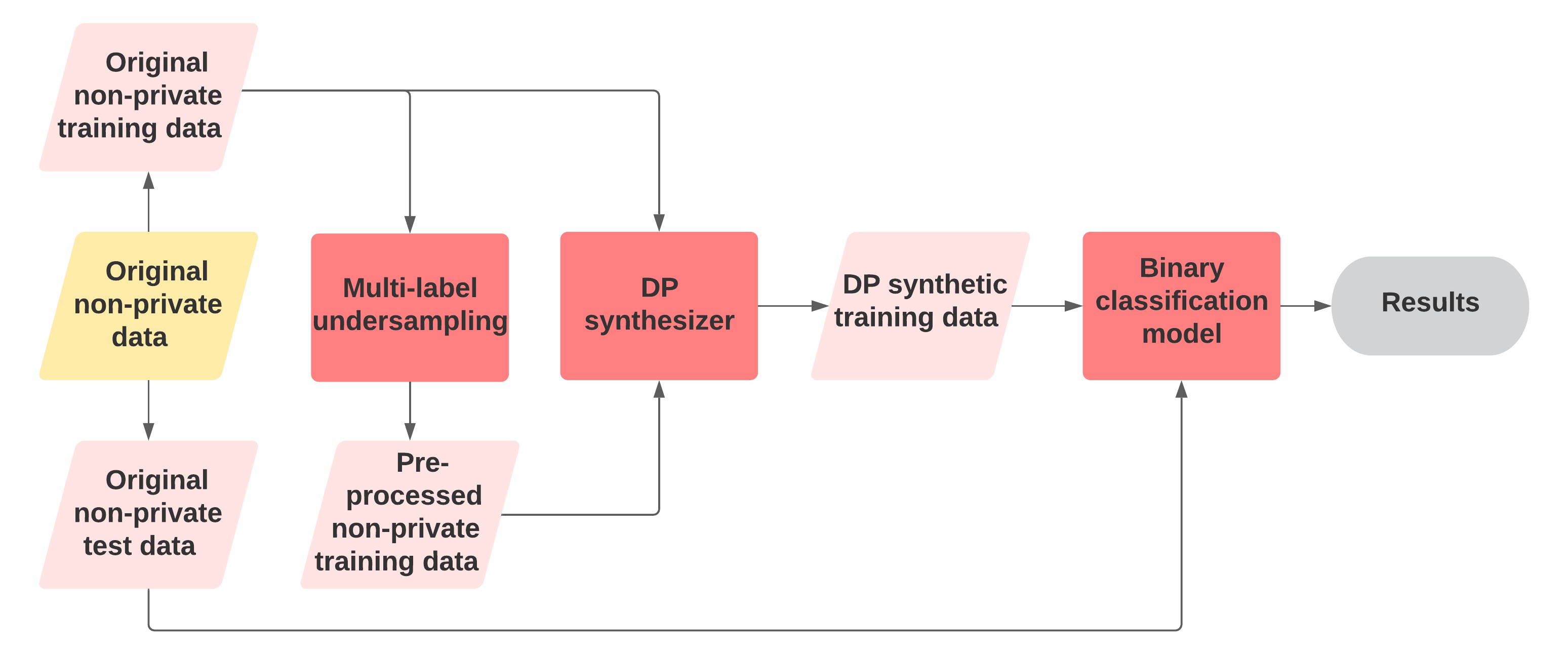}
\caption{Data pipeline used to evaluate the downstream fairness outcomes of classifiers trained on DP synthetic data. Note that separate pipelines were used to train DP synthesizers on the original non-private training data and pre-processed non-private training data.}
\label{fig:data_pipeline}
\end{figure*}

\textbf{Data.} We ran experiments on three data sets, including the ProPublica COMPAS data set \cite{compas} and the UCI Adult data set \cite{uci}. We also used the ACS Income data set, which has been proposed as a substitute for Adult \cite{ding2021retiringadult}. For each data set, we focused on a particular binary classification task (predicting recidivism in COMPAS and income category in Adult and ACS Income), and measured equalized odds distances with respect to a binary protected attribute of interest (race in COMPAS and gender in Adult and ACS Income).

\section{Methodology}

Our results pipeline is summarized in Figure \ref{fig:data_pipeline}. For each synthesizer and data set, we split the original data into train and test sets. This allowed us to train a synthesizer and generate DP synthetic data while holding out some non-private data for evaluation. We then trained logistic regression binary classifiers with $L_2$ regularization on the synthetic data and evaluated their predictions on the original, non-private data. We repeated this procedure across $\varepsilon=[1.0,2.0,\dots,8.0]$. To understand the variability of our results, we performed ten trials at each privacy budget, generating a total of $80$ synthetic data sets.

To mitigate unfairness in downstream classifiers, we repeated these experiments on pre-processed training sets that balanced the number of observations with respect to both the relevant protected attribute $A \in \{0,1\}$ and label $Y \in \{0,1\}$. This simple pre-processing method, which we call ``multi-label undersampling,'' identifies the minority group $\{a,y\} \in A \times Y$ and randomly undersamples observations from the other three groups until the counts of all four are equal. The resulting data set is still non-private but removes pre-existing class imbalances. As discussed below, this encourages synthesizers to generate more balanced data and binary classification models trained on those data to make more fair predictions.

For each data set and synthesizer, therefore, we generated a total of 160 synthetic data sets (80 with pre-processing, and 80 without it). In the following section, we describe the differences among these data sets and the fairness and accuracy of logistic regression classifiers trained on them.

\section{Experimental Results}

\begin{figure*}[t]
\centering
\includegraphics[width=0.8\textwidth]{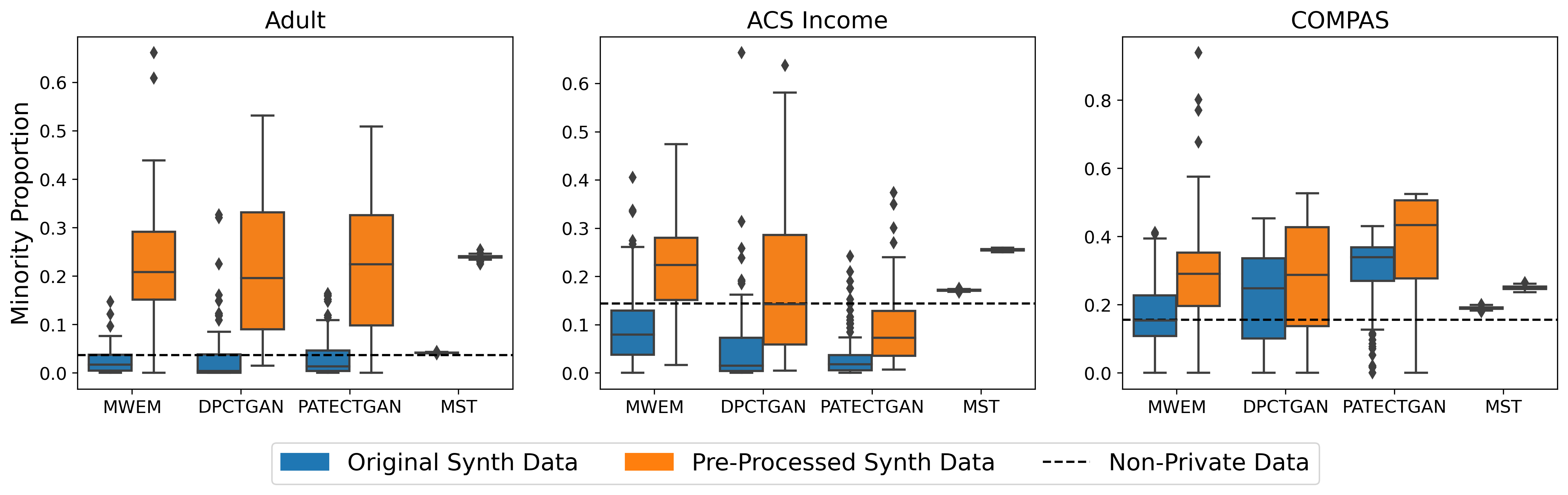}
\caption{Distributions of the proportion of the minority group (with respect to the relevant protected attribute $A$ and label $Y$) in synthetic data generated across data sets and synthesizers. Each blue boxplot corresponds to 80 synthetic data sets generated on the original data, while each orange boxplot shows the results for 80 synthetic data sets obtained by training synthesizers on pre-processed data. The black dotted lines represent the proportion of the minority group present in the original, non-private training data.}
\label{fig:proportions_boxplots}
\end{figure*}

In our experiments, we found that privacy budget did not have a significant impact on the fairness and accuracy of downstream classifiers. More specifically, the variation in the results obtained across trials at fixed epsilon values was generally larger than the differences between them. Therefore, we analyzed the 80 synthetic data sets for each synthesizer and data set in aggregate. We note that restricting the privacy budget to lower $\varepsilon$ values yielded similar results, so we include our results up to $\varepsilon=8.0$ for completeness. 

The differences observed among synthesizers, however, were more significant and are the focus of our discussion below. QUAIL-DPCTGAN and QUAIL-PATECTGAN showed highly variable performance, indicating that hyperparameter tuning may be required to reliably use GAN-based synthesizers in practice, even when they are augmented with QUAIL. 

\subsection{Synthetic Data Distributions}

To understand how DP synthesizers might affect fairness outcomes on downstream classification tasks, we first analyzed the proportion of the minority group $\{a,y\} \in A \times Y$ present in the synthetic data sets. As shown in Figure \ref{fig:proportions_boxplots}, QUAIL-MWEM, QUAIL-DPCTGAN, and QUAIL-PATECTGAN frequently \emph{decreased} this proportion relative to the original, non-private data. This suggests that these synthesizers may exacerbate pre-existing class imbalances in the original data. However, training synthesizers on data that were pre-processed with multi-label undersampling mitigated this issue, yielding synthetic data sets with minority proportions closer to 0.25.

Interestingly, MST does not appear to suffer from this issue and generated synthetic data with minority proportions almost exactly equal to the non-private proportions in Adult, ACS Income, and COMPAS. Further, the variation in the MST minority proportions is significantly smaller than those yielded by the other three synthesizers. These observations indicate that pre-processing the non-private data may not be necessary when using MST.

\subsection{Binary Classification}

Having analyzed the differences among various DP synthetic data sets in terms of their minority group proportions, we turn our attention to how these differences manifest in the fairness and accuracy of downstream logistic regression classifiers.

\begin{figure}[h]
\centering
\includegraphics[width=0.475\textwidth]{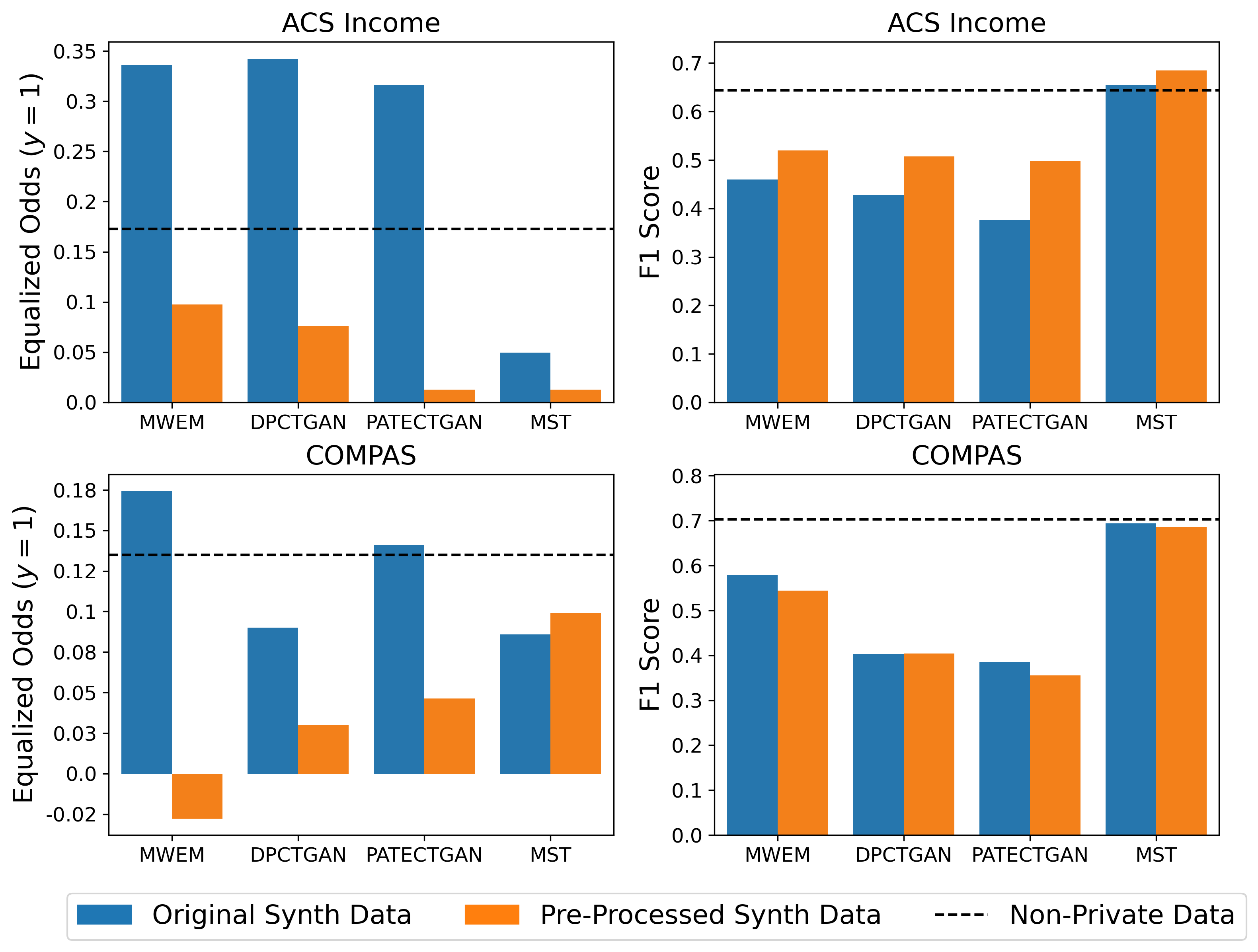}
\caption{Average fairness and accuracy metrics obtained across synthesizers on the ACS Income and COMPAS data sets. Each orange and blue bar indicates the average metric obtained with and without pre-processing, respectively, while the dotted line shows the metric obtained using the original, non-private training data.}
\label{fig:metrics_barcharts}
\end{figure}

Figure \ref{fig:metrics_barcharts} indicates that, on average, classifiers trained on data generated by QUAIL-MWEM, QUAIL-DPCTGAN, and QUAIL-PATECTGAN had higher equalized odds distances (i.e., were less fair) than those trained on non-private training data. This is particularly evident on the ACS Income data, where the average equalized odds distances are nearly twice the non-private metric. Similar results, not shown here, were obtained on the Adult data set.

Further, we observed a strong association between the minority group proportions visualized in Figure \ref{fig:proportions_boxplots} and the downstream fairness outcomes shown in Figure \ref{fig:metrics_barcharts}. In particular, synthetic data sets with lower minority proportions than non-private data (such as those generated on ACS Income) are associated with less fair outcomes, while synthetic data sets that do not decrease this proportion (such as those generated on COMPAS), are less likely to degrade fairness. Intuitively, classifiers trained on data sets containing fewer observations of a particular group are more likely to make incorrect predictions with respect to that group.

The same reasoning can be used to explain the reverse outcome. In particular, our pre-processing method, which encourages synthesizers to generate data with higher proportions of the minority group, is also associated with lower equalized odds distances (more fair outcomes). Similarly, MST did not significantly alter the minority group proportions in comparison to non-private data and therefore did not degrade fairness.

Contrary to the frequently cited trade-off between fairness and accuracy, Figure \ref{fig:metrics_barcharts} also shows that improvements in fairness were granted with virtually no reductions in accuracy. In fact, the most fair and accurate results were both achieved by MST, which had F1-scores nearly equivalent to those obtained on non-private data. The average accuracies achieved by the other three synthesizers, however, were significantly lower.

\section{Conclusion}

In this paper, we studied the effect of differentially private synthetic data on downstream fairness outcomes. We found that three out of the four synthesizers investigated frequently degrade fairness and drew an association between less fair outcomes and decreased proportions of minority groups in the generated synthetic data. This motivated our method of pre-processing the non-private training data, which encouraged synthesizers to generate more balanced classes and mitigated unfair outcomes while retaining predictive accuracy. However, The MST synthesizer achieved fairness and accuracy metrics that were close to those obtained using non-private data -- even without pre-processing -- and may be a preferable option for real-world applications involving DP synthetic data.


\bibliography{example_paper}
\bibliographystyle{icml2020}

\end{document}